%% file: main.tex
\documentclass{article}

    \usepackage[final,nonatbib]{neurips_2021}

\usepackage[utf8]{inputenc} 
\usepackage[T1]{fontenc}    
\usepackage{hyperref}       
\usepackage{url}            
\usepackage{booktabs}       
\usepackage{amsfonts}       
\usepackage{nicefrac}       
\usepackage{microtype}      
\usepackage{xcolor}         

\usepackage{subfigure}
\usepackage{graphics}
\usepackage{graphicx}
\usepackage[inline]{enumitem}
\usepackage{caption} 
\usepackage{longtable}
\usepackage{multirow}
\usepackage{multicol}
\usepackage{fixltx2e}
\usepackage{subfigure}
\usepackage{xcolor}
\usepackage{amsmath}
\mathchardef\mhyphen="2D 

\DeclareRobustCommand{\orderof}{\ensuremath{\mathcal{O}}}

\definecolor{colorA}{rgb}{0.12156862745098039, 0.4666666666666667, 0.7058823529411765}
\definecolor{colorB}{rgb}{1.0, 0.4980392156862745, 0.054901960784313725}
\definecolor{colorC}{rgb}{0.17254901960784313, 0.6274509803921569, 0.17254901960784313}
\definecolor{colorD}{rgb}{0.8392156862745098, 0.15294117647058825, 0.1568627450980392}
\definecolor{colorE}{rgb}{0.5803921568627451, 0.403921568627451, 0.7411764705882353}

\title{Resource Allocation in Disaggregated Data Centre Systems with Reinforcement Learning}

\author{
  Zacharaya Shabka, Georgios Zervas \\
  Optical Networks Group, University College London \\
  \texttt{\{zacharaya.shabka.18, g.zervas\}@ucl.ac.uk} \\
}

\begin{document}

\maketitle

\begin{abstract}

Resource-disaggregated data centres (RDDC) propose a resource-centric, and high-utilisation architecture for data centres (DC), avoiding resource fragmentation and enabling arbitrarily sized resource pools to be allocated to tasks, rather than server-sized ones. RDDCs typically impose greater demand on the network, requiring more infrastructure and increasing cost and power, so new resource allocation algorithms that co-manage both server and networks resources are essential to ensure that allocation is not bottlenecked by the network, and that requests can be served successfully with minimal networking resources. We apply reinforcement learning (RL) to this problem for the first time and show that an RL policy based on graph neural networks can learn resource allocation policies end-to-end that outperform previous hand-engineered heuristics by up to 22.0\%, 42.6\% and 22.6\% for acceptance ratio, CPU and memory utilisation respectively, maintain performance when scaled up to RDDC topologies with $10^2\times$ more nodes than those seen during training and can achieve comparable performance to the best baselines while using $5.3\times$ less network resources.

\end{abstract}

\input{submission/include}
\bibliography{bibliography}
\newpage
\input{submission/appendix}
\end{document}

%% file: submission/include.tex
\input{submission/introduction}
\input{submission/methodology}
\input{submission/experiment}

\input{submission/conclusion}


%% file: submission/introduction.tex
\section{Introduction and related work}

Modern data centre (DC) architectures are and have historically been server-centric, whereby compute resources (CPU, memory etc) are co-located on the same server in small quantities, which are in turn connected via multi-tier electronic/opto-electronic packet switched (EPS) network. Fundamentally, this architecture poses two primary limitations. Firstly, disparity between request sizes and the resources available on physical servers causes resource fragmentation (e.g. if all the CPU and only some of the memory on a server is allocated to a user, the rest of the memory is effectively wasted) \cite{Amaral2021}. Secondly, EPS networks have limited bandwidth per-port, and are built in oversubscribed architectures due to cost limitations. As such, they are not able to provide the performance (bandwidth, latency) or communication determinism for non-local resources to be pooled and allocated to individual tasks with local-like performance.

Resource-disaggregated DC architectures (RDDC) present an alternative DC architecture whereby pools of server resources can be allocated across non-local servers (`virtual data centres' - VDC), enabling a high-utilization resource-centric environment. This would enable DC operators to avoid resource fragmentation and allow users to exploit arbitrarily sized resource pools, rather than being bound to server-sized quantities. VDCs can in principle be treated like regular DC-clusters, whereby arbitrary tasks/schedulers/etc can run on the VDC's resources provided that two primary difficulties can be overcome. Firstly, local-like communication performance and determinism between non-local resources must be guaranteed by the network. Secondly, new allocation algorithms must be developed to co-manage the compute and networking resources. VDC requests require both server and network resources be reserved, so impose greater demand on the RDDCs network which can in turn bottleneck resource allocation. Co-managing compute and network resources so that the amount of network infrastructure needed can be reduced without compromising the resource efficiency or inter-resource communication of the system is essential for network cost, complexity reduction and performance.

Experimental work has shown that networks supporting the communication requirements (high bandwidth, low latency) and determinism needed by RDDCs can be built using commodity hardware. For example, benchmark tasks (e.g. STREAM) run on resource pools disaggregated over circuit-switched optical networks have achieved performance similar to that of server-local pools \cite{Mishra2020, Mishra21, Bielski18}.

Co-allocation of server and network resources maps to an online case of the Quadratic Assignment problem, a canonical combinatorial optimisation problem \cite{Meng2010}. Network aware resource allocation algorithms for RDDC systems show improvements over traditional packing methods \cite{Yuan18,Zervas18,trafpy}, though are recursive and so don't scale effectively to large RDDC topologies. Similarly, multi-dimensional packing methods have shown improvement over typical packing methods in the context of computer cluster resource allocation when both compute and networking resources of nodes are accounted for \cite{Grandl14}, but do not account for the full state of the network including its higher tiers. Moreover, heuristics in general are limited by the ability of their designers to relate RDDC features to high-level goals such as increasing acceptance ratio, which is a difficult task and mandates many assumptions which may not be appropriate. Reinforcement learning (RL) based on graph neural networks (GNN) has been shown to perform very well in various graph-based optimisation problems in computer-system such as task scheduling and placement \cite{addanki2019placeto, Mao2019, Paliwal2020Reinforced}.  well as in network embedding and routing tasks \cite{Yan2020, Suarez2019}. We propose the use of a similar architecture to solve the problem of network aware resource allocation, where we are not aware of any work applying these methods to this problem to date.

 We present a resource allocation method that can account for both compute and network resources to maximise the number of requests that can be successfully served and make better use of the compute resources whilst minimising the amount of networking resources needed in order to achieve this performance.  

We demonstrate a RL-based network aware resource allocation method that
\begin{enumerate*}
    \item scales to real cluster-scale RDDC topologies;
    \item account for the state of higher tiers of the network;
    \item achieves good performance with less resource required in the higher tiers of the network;
    \item learn its own relationships between basic RDDC features and allocation decisions relative to some high level goal.
\end{enumerate*}

\textbf{Contributions:} We define the problem and formulate it as a markov decision process (MDP) in section \ref{section:problem} and describe the RL model used in section \ref{section:model}. The training scheme and test results compared against 4 baseline heuristics are shown in sections \ref{section:topologies} and \ref{section:results} respectively. Open questions and limitations of the work are discussed in section \ref{section:conclusions}.

%% file: submission/methodology.tex
\section{Problem statement and RL model}

\subsection{Problem}
\label{section:problem}


We model the RDDC as a circuit switched network whereby network switches have a finite number of ports (modelled as distinct channels in an incoming or outgoing link) that can be allocated to a single path only. A RDDC is represented by a graph, $G(V,E)$, where the feature vectors $[v_{cpu},v_{mem}] \forall v \in V$ and $[e_{ch}] \forall e \in E$ represent the CPU and memory resource-units at each node and the number of free channels at each link respectively. Requests arrive at the RDDC and are served one at a time and are not seen in advance, where $R = [r_{cpu},r_{mem},r_{t}]$ represents the CPU, memory and holding time requirements for that request. To satisfy a request, a set of nodes, $V^{'}$, must be found (selected iteratively) such that 
\begin{enumerate*}
    \item $\sum v_{x} \in V^{'} \geq r_{x}, x \in \{cpu,mem\}$;
    \item a set of $\frac{|V^{'}|(|V^{'}|-1)}{2}$ distinct paths can be found to guarantee all-to-all connectivity between all $v \in V^{'}$.
\end{enumerate*}
A solution to this problem is considered to be one which can maximise the number of allocations made over time, given a series of $N$ requests.

\subsection{MDP Formulation}
\label{section:mdp}
\textbf{Episode:} An episode concludes when $N$ requests have been received and either successfully or unsuccessfully allocated. \textbf{State:} given an awaiting request $R$ = $\{r_{cpu},r_{mem},r_t\}$ (CPU, memory and holding-time requirements respectively) the MDP's state is $s$ = $\{G(V_{norm},E),r_{t},U_{cpu},U_{mem}\}$, where $V_{norm}$ = $\{\frac{v_{x}}{r_{x}} \forall v_{x} \in V\},x \in \{cpu,mem\}$ and $U_{x}$ is the RDDC-global utilisation of resource $x \in \{cpu,mem\}$. This combines both node-, edge- and graph- level resource information within the state representation.
\textbf{Action:} A server $v \in V$ is added to $V^{'}$. If constraint 2 in section \ref{section:problem} cannot be satisfied (when k=3 shortest paths are tried per node-pair) the allocation fails. If constraint 2 is satisfied, then $max(v_{x},r_{x})$ from $v$ ($x \in \{cpu,mem\}$) and one channel-per-link per-path per-node-pair are allocated. If constraint 1 \& 2 are satisfied the allocation is successful. Per request, actions are taken until an action succeeds or fails an allocation, at which point a new request is fetched (until termination). \textbf{Reward:} An agent will receive $\alpha$ if the action is successful, $\beta$ if the action is unsuccessful, $\gamma$ otherwise where $10$, $-10$ and $0$ were used for $\alpha$, $\beta$ and $\gamma$ respectively. This ensures that the reward is agnostic to the size of the request, since it is desirable to have a fair allocator that does not systematically prioritise some sub-set of all requests.

\subsection{Model}
\label{section:model}
The learning model consists of a GNN (based on the GraphSAGE architecture \cite{Hamilton2017}) and 2 deep neural networks (DNN) which we refer to as DNN\textsubscript{1} and DNN\textsubscript{2}. The GNN acts on $G(V_{norm},E)$ to generate embeddings of each node in the topology. DNN\textsubscript{1} outputs a high dimensional representation of $[r_{t},U_{cpu},U_{mem}]$. DNN\textsubscript{2} then calculates logits for each node in the RDDC graph based on an input of the concatenation of
\begin{enumerate*}
    \item the GNNs embedding of that node,
    \item the output of DNN\textsubscript{1} and
    \item the element-wise mean of the embeddings of the nodes that have already been allocated to that request (or a zero-vector if the request has just been received and nothing has been allocated yet),
\end{enumerate*}
where these logits are passed through a softmax function to specify the probability of choosing each node. This model is used to approximate a policy, trained using the proximal policy optimisation (PPO) RL algorithm, and is implemented using RLlib and Deep Graph Library \cite{rllib,dgl}. Additional details are provided in Appendix \ref{appendix:model_vis}.
We compare our model against 3 baselines (Tetris, NALB, NULB) from previous work as well as a random allocation policy \cite{Grandl14,Yuan18,Zervas18}.

%% file: submission/experiment.tex
\begin{figure}[h!]
    \centering
    \begin{subfigure}{}
    \includegraphics[width=0.3\textwidth]{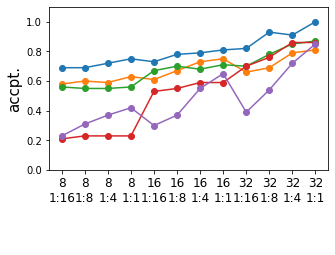}
    \end{subfigure}
    \begin{subfigure}{}
    \includegraphics[width=0.3\textwidth]{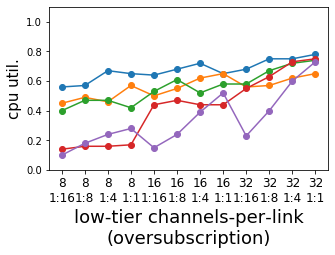}
    \end{subfigure}
    \begin{subfigure}{}
    \includegraphics[width=0.3\textwidth]{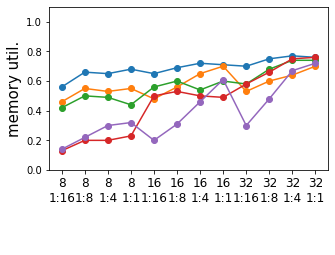}
    \end{subfigure}
    \caption{Comparison of acceptance ratio (left), CPU utilisation (middle) and memory utilisation (right) of the RL agent and all baselines across each [topology,oversubscription-ratio] pair.\newline \textbf{Legend: \textcolor{colorA}{blue}=RL agent, \textcolor{colorB}{orange}=Tetris, \textcolor{colorC}{green}=NALB, \textcolor{colorD}{red}=NULB, \textcolor{colorE}{purple}=random.}}
\label{fig:line}
\end{figure}

\section{Experiment}

\subsection{RDDC topologies and requests}
\label{section:topologies}
Servers initially have 16 units of both CPU and memory resources. We use 12 different 3-tier (typical for DC architectures) topologies, distinguished by the number of ports per-switch-per-server in the lowest tier of the network (i.e. how many distinct channels exist between a server and its rack switch) and the bottom-to-top oversubscription ratio. 
The values for channels-per-link in tier-1 (i.e. max number of other servers that a server connect to) is \{8,16,32\} and the set of bottom-to-top oversubscription ratios is \{1:1,1:4,1:8,1:16\}, where tier-2 and tier-3 channel values are set to ensure these ratios given the number of channels in a tier-1 link (detailed in Appendix \ref{appendix:topologies}). Higher oversubscription (lower ratio) imposes a stronger mandate for rack-locality on allocations due to limited upper-tier network resources. Less tier-1 channels per server limits how many servers an allocation can be spread across, since all server within an allocation must be interconnected. Topologies used for training and testing have 64 nodes (2 clusters $\times$ 2 racks $\times$ 16 servers). Trained models were also tested on graphs the order of $10^2$ times larger with respect to number of nodes (8 clusters $\times$ 8 racks $\times$ 16 servers). Requests are uniformly sampled with a maximum request size of 8 full-servers worth of resources in each domain (128 units), and their holding times are sampled such that the average offered load on the RDDC system is 95\% of all CPU and memory resources. Separate agents are trained for each topology, and tested against each baseline on that same topology.

We evaluate all methods on the basis of three metrics; acceptance ratio, CPU utilisation and memory utilisation. Acceptance ratio refers to the proportion of all requests received by some allocation method that were successfully allocated. CPU and memory utilisation refers to the proportion of the total amount of that resource that is available in the RDDC which is currently allocated to some request. 
\subsection{Results}
\label{section:results}

\begin{table}
\caption{Percentage improvement of the agent pair over the second best performing baseline for that topology across all tested topologies.}
\label{table:comparison}
\begin{tabular}{llllllllllllll}
\multicolumn{14}{c}{\textbf{RL agent improvement over best baseline (\%)}} \\
\toprule
\multicolumn{14}{c}{\textbf{Oversubscription}} \\
\midrule
{} & {} &          \multicolumn{3}{c}{1:16} &          \multicolumn{3}{c}{1:8} &          \multicolumn{3}{c}{1:4} &          \multicolumn{3}{c}{1:1} \\
{} & {} & \multicolumn{3}{c}{accpt, cpu, mem} & \multicolumn{3}{c}{accpt, cpu, mem} & \multicolumn{3}{c}{accpt, cpu, mem} & \multicolumn{3}{c}{accpt, cpu, mem} \\
\midrule
 \textbf{Low-tier} & 8.0  & \multicolumn{3}{c}{ 19.0, 24.4, 21.7} & \multicolumn{3}{c}{ 15.0, 16.3, 20.0} & \multicolumn{3}{c}{ 22.0, 42.6, 22.6} & \multicolumn{3}{c}{ 19.0, 14.0, 23.} \\
 \textbf{channels} & 16.0 & \multicolumn{3}{c}{  9.0, 20.8, 16.1}  & \multicolumn{3}{c}{ 11.4, 11.5, 15.0}  & \multicolumn{3}{c}{  8.2, 16.1, 10.8}  & \multicolumn{3}{c}{    8.0, 0.0, 1.} \\
 \textbf{per-link} & 32.0 & \multicolumn{3}{c}{ 17.1, 17.2, 20.7} & \multicolumn{3}{c}{ 19.2, 11.9, 10.3} & \multicolumn{3}{c}{    5.8, 2.7, 2.7} & \multicolumn{3}{c}{   14.9, 4.0, 0.} \\
\bottomrule
\end{tabular}
\end{table}

Table \ref{table:comparison} shows the percentage improvement that the agent achieved compared to the best performing baseline for acceptance ratio, cpu- and memory- utilisation for all 12 topologies (raw results in Appendix \ref{appendix:table}), where (best,worst) improvement is (22.0\%,5.8\%), (42.6\%,0\%) and (23\%,0\%) for acceptance, CPU and memory utilisation respectively. Fig. \ref{fig:line}, plotting the average performance for acceptance, CPU and memory utilisation per-method per-topology, shows that while the agent is consistently the best performing method, no single baseline is consistently better than the others. The agent is able to learn policies relevant to the specific topology at hand, rather than implement a single static policy (as is done with heuristic methods) that may not be equally effective in different scenarios.
Crucially, given a set number of channels per tier-1 link in the network, the agent is able to achieve similar performance (+9\% best case, -5\% worst case) at a 1:16 oversubscription ratio than the best performing baseline achieves at 1:1 oversubsctiption ratio, reducing the required networking resources in the upper tiers of the RDDC system by $5.3\times$ with respect to the number of ports used, without significant performance degradation (elaborated in Appendix \ref{appendix:resources}). The agent also maintains performance on RDDC topologies with $10^2\times$ more nodes than those seen during training; 16\% better and 3\% worse acceptance in the best and worst case compared against its performance on the smaller topology. For RDDC topologies with $|V|$ nodes and $|E|$ edges, the runtime of the RL agent and Tetris is $\orderof(|V|)$ per action. For the NALB and NULB methods, which are based on a breadth first search procedure, $\orderof(|V| + |E|)$ steps are required per action. Random selection requires $\orderof(1)$ steps per action, but is trivial. The RL agent and Tetris are both linear with respect to the size (number of nodes) of the RDDC. In general, $|E| = f(|V|)$ given the rules of some topology structure and a number of nodes (e.g. $|E| = \frac{(|V|-1)^{2}}{2}$ for all-to-all networks), so runtime of the NALB and NULB methods will generally scale super-linearly with respect to the size of the RDDC.


\begin{figure}[h!]
    \centering
    \includegraphics[width=\textwidth]{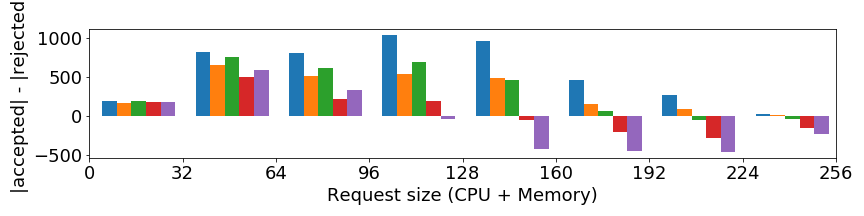}
    \caption{Histogram of the difference between the number of accepted and rejected requests relative to the number of CPU + memory units requested, shown across results from all tests combined.\newline \textbf{Legend: \textcolor{colorA}{blue}=RL agent, \textcolor{colorB}{orange}=Tetris, \textcolor{colorC}{green}=NALB, \textcolor{colorD}{red}=NULB, \textcolor{colorE}{purple}=random.}}
    \label{fig:success}
\end{figure}

Figure \ref{fig:success} shows the delta between the number of accepted and rejected requests of each method across all topologies against the total size of the request (requested CPU + memory units). The RL-based method is advantageous across the full range of request sizes. Most notably, in the upper 50\% of request sizes, the agent always maintains a positive delta. On the other hand, all baselines except Tetris fall into a negative delta at various regions within the upper half of request sizes, where NULB (and random) have a negative delta throughout the entire region. This shows that the RL-agent is superior across the entire range of request sizes, rather than being strongly biased in favour of particular ranges.

%% file: submission/conclusion.tex
\section{Discussion and future work}
\label{section:conclusions}

In this work we show that RL-based co-allocation of compute and network resources, in the context of resource disaggregated data centre (RDDC) systems, can outperform several hand-engineered baselines with respect to acceptance ratio, CPU utilisation and memory utilisation on the same topologies, achieve similar performance on topologies with $5.3\times$ less networking resources (reducing the cost and power consumption of the RDDC), and maintain performance when scaled up to RDDC topologies $10^{2}\times$ bigger than those seen during training.

Future work will examine more complex request representations (e.g. non-uniform distributions that can change over time); different disaggregated resource structures where not all servers contain all types of resources; different reward structures to explore how different behaviours are incentivised, such as rewards based on server utilisation rather than acceptance.

Finally, we will also seek to probe the behaviour of all methods more deeply than by evaluation on the basis of the simple metrics used here. This will allow for deeper understanding of the nature of the policies encoded in the baselines and learnt by the RL agent, allowing for a more robust analysis of how or why the RL agent is able to achieve better performance and from which aspects of the baseline methods do their bottlenecks arise.

\paragraph{Acknowledgments.} This work was supported under the Engineering and Physical Sciences Research Council (EP/R041792/1 and EP/L015455/1), the Industrial Cooperative Awards in Science and Technology (EP/R513143/1), the OptoCloud (EP/T026081/1), and the TRANSNET (EP/R035342/1) grants.

%% file: submission/appendix.tex
\appendix
\section{Model details}
\label{appendix:model_vis}
\begin{figure}[h!]
    \centering
    \includegraphics[width=\textwidth]{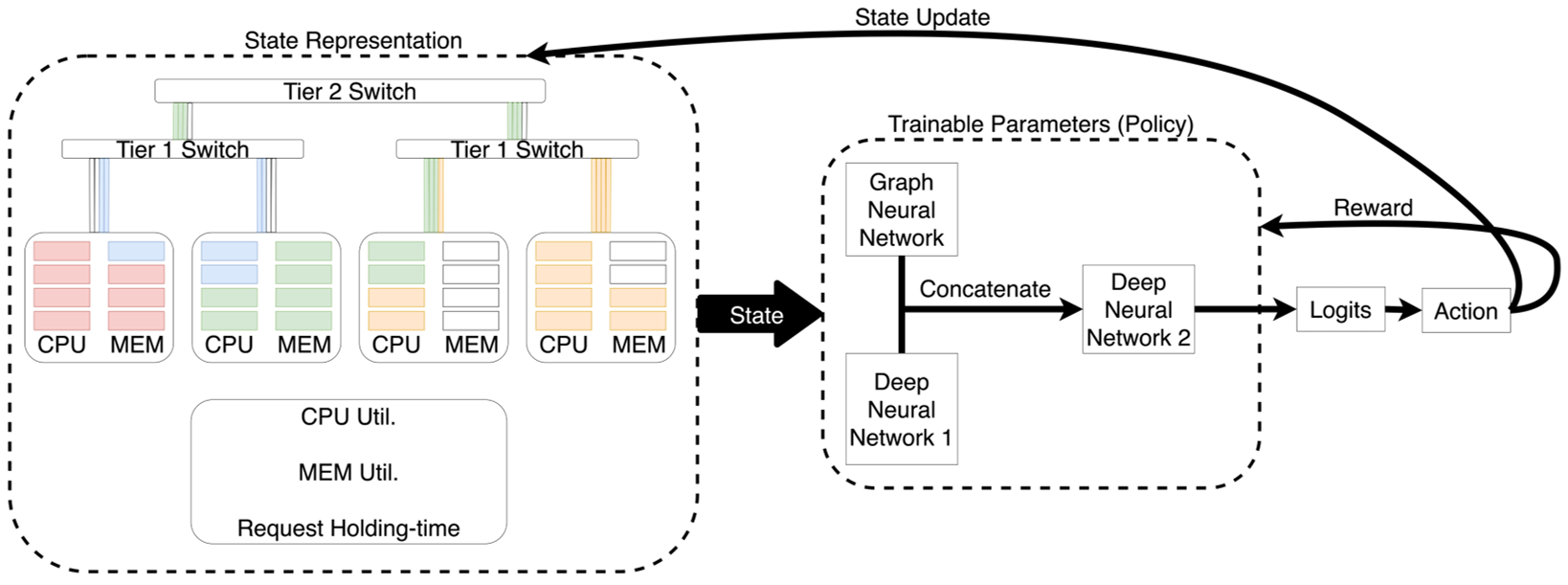}
    \caption{High-level diagram of the RDDC + model + RL feedback loop implemented in this work.}
    \label{fig:model}
\end{figure}

The GNN used a distinct mean-based aggregator for each layer, where messages exchanged during the message passing process are aggregated like:
\begin{equation}
    v = \frac{1}{|N(v)|+1} \sum_{x \in M_{v}} W_{i}(x)
\end{equation}
where $v$ is a node (embedding) in the RDDC graph $G(V,E)$, $N(v)$ is the set of the one-hop neighbours of $v$, $M_v$ is the set of messages received by $v$ from it's neighbours and $W_i$ is a neural network associated with the $i^{th}$ layer of the GNN.

The GNN outputs embeddings of each node in 16 dimensions, and 3 layers were used so that information from the top of the network can be accounted for in the embeddings of the servers.

\section{Calculating network resource efficiency}
\label{appendix:resources}

When we refer to one topology having less networking resources in the upper tiers of the network than another, we refer to the comparison of two topologies with the same number of channels-per-link in the lowest tier (i.e. the same amount of networking resources directly connecting servers to their rack switch) but different oversubscription ratios.

For example, the topology in Appendix \ref{appendix:topology_vis} has 8 links connecting tier-1 and tier-2, and 4 links connecting tier-2 and tier-3. In the version of this topology when the server-rack links have 8 channels each, the tier-1 - tier-2 links have 16 channels and the tier-2 - tier-3 links have 4 in the highest oversubscription case (1:16). In the lowest oversubscription case (1:1), on the other hand, they have 64 and 64 respectively. This means that the total number of channels in the higher tiers in the 1:1 oversubscription case is $\frac{16}{3} \approx 5.3$ greater than in the 1:16 oversubscription case.

The full results in Appendix \ref{appendix:table} shows that the acceptance ratio scores achieved by the agent in the 1:16 oversubscription are worst case -6\% and best case +10\% compared to the best performing baseline in the topology with 1:1 oversubscription but the same number of channels per link between servers and racks.

\section{Topology visualisation}
\label{appendix:topology_vis}

\begin{figure}[h!]
    \centering
    \includegraphics[width=0.5\textwidth]{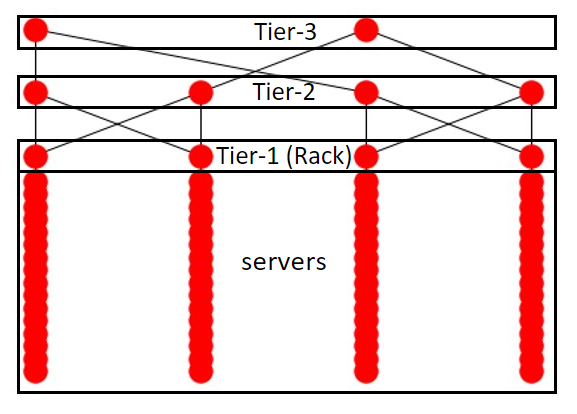}
    \caption{Visualisation of the (smaller) topology used for training and testing in this work.}
    \label{fig:topology}
\end{figure}

Figure \ref{fig:topology} shows the graph-structure of the smaller RDDC used during training. Note that this topology, each server (lower vertical groups of nodes) are each directly connected to their corresponding rack switch (3rd row of switches from the top) with a unique link.

The RDDCs used in this work were simulated, and their topology was designed on the basis of the Fabric \cite{FacebookFabric}. The network has 3 tiers of switches, where only rack switches (the lowest tier) are connected directly to servers. Switches in the other tiers connect only to other switches in higher and lower tiers. Each link in the topology is described as having multiple unique `channels'; this is equivalent (in the circuit switching paradigm explored here) to each switch having a distinct number of unique ports.

\section{Table of topology oversubscription structures}
\label{appendix:topologies}

Table \ref{table:oversub} shows how the oversubscription structure was determined given a top-to-bottom oversubscription and a number of channels per tier-1 link. The `oversub' value for each tier represents the oversubscription at the interface between a switch at that tier and the tier below it.

\begin{table}[h!]
\centering
\caption{Oversubscription and number of channels per link for each topology. `Bottom-top oversubscription' refers to the oversubscription from the servers to the top tier of switches (tier-3). `Oversub' refers to the oversubscription at the interface between that tier and the tier below it (hence Tier-1 does not have an `oversub' value. In this work we used topologies of this structure with $n \in \{8, 16, 32\}$.}
\label{table:oversub}
\begin{tabular}{lllllllll}
\multicolumn{9}{c}{\textbf{Oversubscription and channels-per-link for higher RDDC topology tiers}} \\
\toprule
\multicolumn{9}{c}{\textbf{Bottom-top Oversubscription}} \\
\midrule
{} & \multicolumn{2}{c}{1:16} &          \multicolumn{2}{c}{1:8} &          \multicolumn{2}{c}{1:4} &          \multicolumn{2}{c}{1:1} \\
{} & \multicolumn{2}{c}{Oversub, Channels} & \multicolumn{2}{c}{Oversub, Channels} & \multicolumn{2}{c}{Oversub, Channels} & \multicolumn{2}{c}{Oversub, Channels} \\
\midrule
Tier-1 & \multicolumn{2}{c}{-, $n$} & \multicolumn{2}{c}{-, $n$} & \multicolumn{2}{c}{-, $n$} & \multicolumn{2}{c}{-, $n$} \\
Tier-2 & \multicolumn{2}{c}{1:4, $2n$} & \multicolumn{2}{c}{1:2, $4n$} & \multicolumn{2}{c}{1:2, $4n$} & \multicolumn{2}{c}{1:1, $8n$} \\
Tier-3 & \multicolumn{2}{c}{1:4, $\frac{1}{2}n$} & \multicolumn{2}{c}{1:4, $n$} & \multicolumn{2}{c}{1:2, $2n$} & \multicolumn{2}{c}{1:1, $8n$} \\
\bottomrule
\end{tabular}
\end{table}

\section{Full table of results}
\label{appendix:table}

\begin{longtable}{llllllll}
\toprule
 Method &          Topology & Accpt. &   CPU &   Mem & SR-Link & RA-Link & AC-Link \\
\midrule
  agent &      8.0\_16.0\_4.0 &   0.69 &  0.56 &  0.56 &    0.55 &    0.95 &    0.79 \\
 tetris &      8.0\_16.0\_4.0 &   0.58 &  0.45 &  0.46 &    0.66 &    0.81 &    0.50 \\
  nalb &      8.0\_16.0\_4.0 &   0.56 &  0.40 &  0.42 &    0.67 &    0.99 &    0.93 \\
  nulb &      8.0\_16.0\_4.0 &   0.21 &  0.14 &  0.13 &    0.89 &    1.00 &    1.00 \\
 random &      8.0\_16.0\_4.0 &   0.23 &  0.10 &  0.14 &    0.94 &    0.84 &    0.24 \\
\midrule
  agent &      8.0\_32.0\_8.0 &   0.69 &  0.57 &  0.66 &    0.51 &    0.93 &    0.56 \\
 tetris &      8.0\_32.0\_8.0 &   0.60 &  0.49 &  0.55 &    0.59 &    0.80 &    0.29 \\
  nalb &      8.0\_32.0\_8.0 &   0.55 &  0.47 &  0.50 &    0.61 &    0.99 &    0.92 \\
  nulb &      8.0\_32.0\_8.0 &   0.23 &  0.16 &  0.20 &    0.86 &    1.00 &    0.99 \\
 random &      8.0\_32.0\_8.0 &   0.31 &  0.18 &  0.22 &    0.87 &    0.82 &    0.21 \\
\midrule
  agent &     8.0\_32.0\_16.0 &   0.72 &  0.67 &  0.65 &    0.43 &    0.88 &    0.72 \\
 tetris &     8.0\_32.0\_16.0 &   0.59 &  0.46 &  0.53 &    0.60 &    0.72 &    0.49 \\
  nalb &     8.0\_32.0\_16.0 &   0.55 &  0.47 &  0.49 &    0.61 &    0.98 &    0.93 \\
  nulb &     8.0\_32.0\_16.0 &   0.23 &  0.16 &  0.20 &    0.86 &    1.00 &    0.99 \\
 random &     8.0\_32.0\_16.0 &   0.37 &  0.24 &  0.30 &    0.79 &    0.68 &    0.21 \\
\midrule
  agent &     8.0\_64.0\_64.0 &   0.75 &  0.65 &  0.68 &    0.44 &    0.55 &    0.39 \\
 tetris &     8.0\_64.0\_64.0 &   0.63 &  0.57 &  0.55 &    0.53 &    0.80 &    0.77 \\
  nalb &     8.0\_64.0\_64.0 &   0.56 &  0.42 &  0.44 &    0.66 &    0.99 &    0.99 \\
  nulb &     8.0\_64.0\_64.0 &   0.23 &  0.17 &  0.23 &    0.84 &    1.00 &    1.00 \\
 random &     8.0\_64.0\_64.0 &   0.42 &  0.28 &  0.32 &    0.78 &    0.83 &    0.77 \\
\midrule
  agent &     16.0\_32.0\_8.0 &   0.73 &  0.64 &  0.65 &    0.68 &    0.90 &    0.80 \\
 tetris &     16.0\_32.0\_8.0 &   0.61 &  0.50 &  0.48 &    0.79 &    0.75 &    0.26 \\
  nalb &     16.0\_32.0\_8.0 &   0.67 &  0.53 &  0.56 &    0.73 &    0.94 &    0.54 \\
  nulb &     16.0\_32.0\_8.0 &   0.53 &  0.44 &  0.50 &    0.73 &    0.95 &    0.59 \\
 random &     16.0\_32.0\_8.0 &   0.30 &  0.15 &  0.20 &    0.94 &    0.83 &    0.19 \\
\midrule
  agent &    16.0\_64.0\_16.0 &   0.78 &  0.68 &  0.69 &    0.66 &    0.94 &    0.75 \\
 tetris &    16.0\_64.0\_16.0 &   0.67 &  0.55 &  0.56 &    0.76 &    0.80 &    0.23 \\
  nalb &    16.0\_64.0\_16.0 &   0.70 &  0.61 &  0.60 &    0.67 &    0.94 &    0.52 \\
  nulb &    16.0\_64.0\_16.0 &   0.55 &  0.47 &  0.53 &    0.70 &    0.95 &    0.62 \\
 random &    16.0\_64.0\_16.0 &   0.37 &  0.24 &  0.31 &    0.89 &    0.84 &    0.13 \\
\midrule
  agent &    16.0\_64.0\_32.0 &   0.79 &  0.72 &  0.72 &    0.64 &    0.85 &    0.73 \\
 tetris &    16.0\_64.0\_32.0 &   0.73 &  0.62 &  0.65 &    0.71 &    0.68 &    0.22 \\
  nalb &    16.0\_64.0\_32.0 &   0.68 &  0.52 &  0.54 &    0.70 &    0.90 &    0.64 \\
  nulb &    16.0\_64.0\_32.0 &   0.59 &  0.44 &  0.50 &    0.74 &    0.93 &    0.74 \\
 random &    16.0\_64.0\_32.0 &   0.55 &  0.39 &  0.46 &    0.78 &    0.66 &    0.12 \\
\midrule
  agent &  16.0\_128.0\_128.0 &   0.81 &  0.65 &  0.71 &    0.66 &    0.90 &    0.85 \\
 tetris &  16.0\_128.0\_128.0 &   0.75 &  0.65 &  0.70 &    0.64 &    0.80 &    0.74 \\
  nalb &  16.0\_128.0\_128.0 &   0.71 &  0.58 &  0.60 &    0.66 &    0.93 &    0.87 \\
  nulb &  16.0\_128.0\_128.0 &   0.59 &  0.44 &  0.49 &    0.73 &    0.97 &    0.94 \\
 random &  16.0\_128.0\_128.0 &   0.65 &  0.52 &  0.61 &    0.66 &    0.74 &    0.65 \\
\midrule
  agent &    32.0\_64.0\_16.0 &   0.82 &  0.68 &  0.70 &    0.76 &    0.82 &    0.44 \\
 tetris &    32.0\_64.0\_16.0 &   0.66 &  0.56 &  0.53 &    0.87 &    0.80 &    0.21 \\
  nalb &    32.0\_64.0\_16.0 &   0.70 &  0.58 &  0.58 &    0.83 &    0.92 &    0.37 \\
  nulb &    32.0\_64.0\_16.0 &   0.70 &  0.55 &  0.58 &    0.80 &    0.88 &    0.31 \\
 random &    32.0\_64.0\_16.0 &   0.39 &  0.23 &  0.30 &    0.95 &    0.84 &    0.17 \\
\midrule
  agent &   32.0\_128.0\_32.0 &   0.93 &  0.75 &  0.75 &    0.76 &    0.91 &    0.58 \\
 tetris &   32.0\_128.0\_32.0 &   0.69 &  0.57 &  0.60 &    0.84 &    0.83 &    0.28 \\
  nalb &   32.0\_128.0\_32.0 &   0.78 &  0.67 &  0.68 &    0.77 &    0.88 &    0.25 \\
  nulb &   32.0\_128.0\_32.0 &   0.76 &  0.63 &  0.66 &    0.73 &    0.91 &    0.33 \\
 random &   32.0\_128.0\_32.0 &   0.54 &  0.40 &  0.48 &    0.89 &    0.84 &    0.17 \\
\midrule
  agent &   32.0\_128.0\_64.0 &   0.91 &  0.75 &  0.77 &    0.82 &    0.69 &    0.18 \\
 tetris &   32.0\_128.0\_64.0 &   0.79 &  0.62 &  0.64 &    0.76 &    0.71 &    0.32 \\
  nalb &   32.0\_128.0\_64.0 &   0.85 &  0.72 &  0.74 &    0.69 &    0.79 &    0.31 \\
  nulb &   32.0\_128.0\_64.0 &   0.86 &  0.73 &  0.75 &    0.68 &    0.77 &    0.32 \\
 random &   32.0\_128.0\_64.0 &   0.72 &  0.60 &  0.67 &    0.78 &    0.66 &    0.11 \\
\midrule
  agent &  32.0\_256.0\_256.0 &   1.00 &  0.78 &  0.76 &    0.80 &    0.83 &    0.77 \\
 tetris &  32.0\_256.0\_256.0 &   0.81 &  0.65 &  0.70 &    0.73 &    0.84 &    0.81 \\
  nalb &  32.0\_256.0\_256.0 &   0.87 &  0.74 &  0.74 &    0.67 &    0.87 &    0.82 \\
  nulb &  32.0\_256.0\_256.0 &   0.86 &  0.75 &  0.76 &    0.68 &    0.87 &    0.80 \\
 random &  32.0\_256.0\_256.0 &   0.85 &  0.73 &  0.72 &    0.65 &    0.73 &    0.64 \\
\midrule
\bottomrule
\label{table:full}
\end{longtable}